# Safe Hierarchical Reinforcement Learning for CubeSat Task Scheduling Based on Energy Consumption

Mahya Ramezani, M. Amin Alandihallaj, Jose Luis Sanchez-Lopez, and Andreas Hein

*Abstract*— This paper presents a Hierarchical Reinforcement Learning (HierRL) methodology tailored for optimizing CubeSat task scheduling in Low Earth Orbits (LEO). Incorporating a high-level policy for global task distribution and a low-level policy for real-time adaptations as a safety mechanism, our approach integrates the Similarity Attention-based Encoder (SABE) for task prioritization and an MLP estimator for energy consumption forecasting. Integrating this mechanism creates a safe and fault-tolerant system for CubeSat task scheduling. Simulation results validate the HierRL's superior convergence and task success rate, outperforming both the MADDPG model and traditional random scheduling across multiple CubeSat configurations.

## I. INTRODUCTION

CubeSats have transformed the space industry, providing a cost-effective and efficient way to conduct diverse space missions, from scientific observations to advanced communications [1, 2]. A rising focus is on equipping spacecraft with advanced autonomous decision-making capabilities [3, 4]. Achieving this relies on using automated planning tools to reduce human involvement and effectively handle complex and uncertain environments. Implementing on-board planning mechanisms in spacecraft missions brings substantial benefits, including increased spacecraft availability, heightened reliability, and reduced ground segment operational costs. However, despite their potential, CubeSats face significant task scheduling challenges in distributed systems due to processing limitations [5]. Efficient energy management is a primary concern, given their reliance on limited solar panel-derived energy. Ensuring they operate within these constraints while maintaining high reliability in space underscores the importance of fault tolerance in satellite operations [6].

In CubeSat operations, the criticality of energy management is accentuated by their inherent power limitations [7]. The complexity of the energy consumption issue is compounded by task-dependent variability, especially in observation missions with sophisticated sensor payloads like high-resolution cameras, adaptive sampling, and data transmission.

M. Ramezani is with the Automation and Robotics Research Group (ARG), Interdisciplinary Centre for Security, Reliability and Trust (SnT), University of Luxembourg (UL) (corresponding author; e-mail: mahya.ramezani@uni.lu).

M. A. Alandihallaj is with the Space Systems Research Group (SpaSys), SnT, UL (e-mail: amin.hallaj@uni.lu).

J. L. Sanchez-Lopez is with the ARG, SnT, UL (e-mail: joseluis.sanchezlopez@uni.lu).

A. Hein is with the SpaSys, SnT, UL (e-mail: andreas.hein@uni.lu).

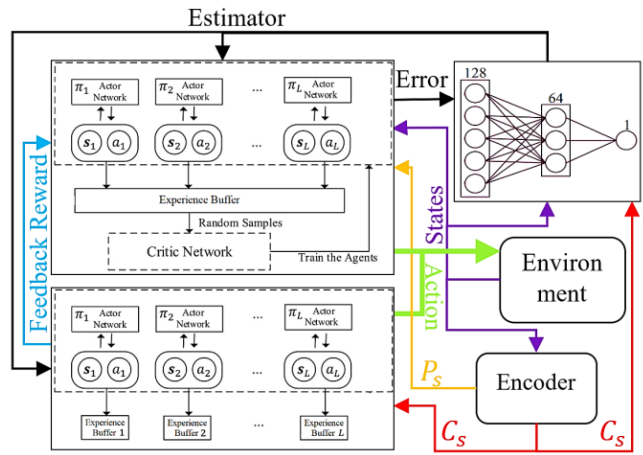

Figure 1 The scematic diagram of the algorithm.

Solving planning problems in this context typically involves a constrained optimization [8]. However, the inherent uncertainties and complexities of space environments, combined with task variability and unpredictability, often surpass the capabilities of traditional tools [9].

One promising solution gaining attention involves applying artificial intelligence to dynamic task scheduling [10]. Artificial intelligence benefits from declining computational costs, abundant data, and advanced algorithms, with Deep Learning (DL) and Reinforcement Learning (RL) playing pivotal roles [11]. RL, a subset of Machine Learning (ML), focuses on training agents to make sequential decisions by interacting with an environment to maximize cumulative rewards [12].

RL has emerged as a crucial paradigm in ML with diverse applications and untapped potential across various domains. One promising application is dynamic task scheduling, where RL algorithms offer significant advantages. The repetitive nature of scheduling decisions aligns well with the data-intensive training methods of RL [13]. Moreover, RL's unique feature is its ability to adaptively make decisions in real-time without requiring a comprehensive environmental model [14, 15].

In the literature, the satellite task allocation field's primary focus is on Earth observation (EO) missions, which present a classic example of a challenging multi-objective combinatorial problem [16]. This complexity makes them suitable candidates for solutions using Deep Reinforcement Learning (DRL) methods. Huang, et al. [10] formulate EO task scheduling problems by introducing the concepts of visible window (VW) and observation window (OW). The decision variables in this context typically involve continuous parameters specifying the start time of OWs for specific targets, alongside binary variables indicating whether an observation task is scheduled or not. The primary objective of EO task scheduling is to maximize observation profit while respecting constraints.

Haijiao, et al. [17] introduce dynamic real-time scheduling for image satellites. Here, observation tasks arrive dynamically with associated rewards, and each task is accepted or rejected based on a policy, provided that on-board data storage and timing constraints are satisfied. This dynamic scheduling problem is formulated as a Dynamic and Stochastic Knapsack Problem [18] and tackled using DRL techniques.

The interaction between attitude changes and observation tasks introduces a time-dependent scheduling challenge, surpassing the complexity of EO systems [19]. In [20], the optimization goals involve maximizing total observation profit while ensuring image quality, which inherently conflict as scheduling more targets can boost profit but potentially compromise image quality. To address this, a two-phase algorithm is proposed. In the first phase, a recurrent neural network learns a scheduling policy for selecting tasks, while in the second phase, a Deep Deterministic Policy Gradient (DDPG) algorithm optimizes the choice of OWs to enhance image quality. A more efficient approach by Wei, et al. [21] introduces a dual objective, considering both the failure rate of observation requests and the timeliness of scheduled requests. Simulation experiments reveal superior performance and faster training, making it suitable for quicker re-training if needed.

The EO scheduling problem becomes notably more intricate when multi-satellite systems, such as satellite constellations, are taken into account. Chong, et al. [22] address multi-satellite cooperative task scheduling using RL. The approach considers a set of satellites, each possessing varying on-board resource capabilities. The approach's objective is to assign each task to a specific satellite based on their available individual resources and the required resources of each task. To enhance the cooperative RL policies of each satellite, a hybrid learning approach that incorporates genetic algorithms is developed. While this approach demonstrates commendable performance in terms of cooperative optimization, it does have limitations. Specifically, it determines task acceptance without specifying OWs for task execution and it assumes complete knowledge of the resource requirements for each task, rendering it impractical for real-world implementation.

Traditional reinforcement learning algorithms struggle with the complexity and scalability challenges inherent in multi-agent task scheduling. Moreover, fully decentralized approaches suffer from computational inefficiency, slow convergence, and demand consistent inter-agent communication, which may not be feasible due to energy and bandwidth limitations in CubeSat networks. On the other hand, Hierarchical Reinforcement Learning (HierRL) addresses these challenges by breaking the problem into high-level and low-level decision-making. This bifurcation expedites convergence, enhances robustness against failures, and allows for more efficient computational resource utilization [23] making it a compelling choice for task scheduling in CubeSat swarms. Such a structure can potentially combine the strengths of traditional algorithms with the adaptability of RL and DRL.

In this paper, we introduce a novel methodology, shown in Fig. 1, that leverages the power of hierarchical deep reinforcement learning for task scheduling in CubeSats. Our primary contributions include the development of a multilayer perceptron (MLP) estimator for task energy consumption that aids both the low-level algorithm and the high-level algorithm in decision-making. We also present an attention-based encoder for scoring tasks, ensuring that the most critical tasks are addressed first. Our hierarchical decision-making structure with high-level and low-level policies act as a safety mechanism for algorithm, optimizes decisions at multiple levels. The detailed reward structure promotes energy consumption, spatial considerations, and deadline adherence. These advancements ensure robust satellite operations, fault-tolerant, and enhanced task prioritization also helps algorithms to faster convergence in essence. The methodology pioneers an energy-conscious perspective in CubeSat task scheduling, fulfilling an evident need in the field.

## II. PROBLEM DEFINITION

We examine a network of CubeSats, denoted as $C_1, C_2, \ldots, C_N$, operating in the Low Earth Orbits (LEO). It is assumed that they are in communication with a ground station, which receive tasks from. Once relayed, these tasks are aggregated within a centralized shared storage, which is universally accessible to each CubeSat constituting the network.

Each CubeSat, $C_i$, is provisioned with a finite energy storage. Energy is depleted during task execution and any ancillary orbital adjustments. Insufficient energy reserves relegate the CubeSat to a dormant mode until replenishment occurs. The rate of energy recovery is primarily influenced by CubeSat's solar panel exposure to sunlight, which itself is a function of its orbital position and diurnal cycles, thereby imparting a cyclical pattern to energy availability.

The energy replenishment rate $E_{r_i}(t)$ for $C_i$ over time $t$ can be mathematically encapsulated as follows:

$$E_{r_i}(t) = \eta(t) A I_{sc} \cos(\theta_{s_t}) \quad (1)$$

where $\eta(t)$ is the efficiency of the solar panels over time, ranging between 0 and 1, $A$ is the effective area of the solar panels exposed to sunlight, $I_{sc}$ is the solar constant, which is approximately $1361\ W/m^2$ for LEO [24] but may vary with solar activity, $\theta_{s_t}$ is the angle between the solar rays and the normal to the plane of the solar panel at time $t$.

The dynamics of the energy storage $E_i(t)$ can thus be expressed as:

$$\dot{E}_i(t) = E_{r_i}(t) - E_{consumed_i}(t) \quad (2)$$

in which $E_{consumed_i}$ represents the energy consumed for task execution and other activities at time $t$.

By solving this differential equation subject to the initial condition, the CubeSat's energy level at any time $t$ can be obtained.

The computational prowess of each CubeSat is principally characterized by its processor capabilities. Each observational task stipulates a unique set of computational prerequisites and a stringent deadline.

The system maintains a central shared storage of tasks accessible to all CubeSats, each defined by attributes including priority from the ground station, computational requirements, OW, and spatial coordinates. Each task is exclusive to one CubeSat. A uniqueness constraint for tasks applied to prevent redundancy and real-time task status updates (*unassigned*, *in progress*, *completed*) ensure constraint satisfaction. The primary goal is to develop a scheduling algorithm optimizing a composite reward function, considering task priority, deadline adherence, resource efficiency, and CubeSat positioning. The algorithm should adapt to dynamic task arrivals, ensuring system flexibility. The objectives are achieved using a Markov game model approach [25].

## III. METHODOLOGY

We present a HierRL method for the scheduling problem for CubeSats considering energy consumption.

In the method, an encoder first prioritized tasks, emphasizing critical attributes that influence the success rate in task scheduling and energy consumption like task duration, spatial constraints, memory, and computational requirements. With the aid of an attention mechanism, the encoder assigns scores for prioritizing each task to speed up the convergence of high-level reinforcement learning.

We introduce an MLP-based estimator that uses the encoder's output and task score. By employing similarity-attention-based mechanisms and referencing data from prior task executions stored in the high-level reinforcement learning replay experience, this estimator predicts the energy consumption for each task based on each task ID.

Subsequently, tasks are allocated to individual CubeSats based on system states and complexity scores through a high-level reinforcement learning process. If a CubeSat is in a failure status or lacks adequate battery power or time for a task, a low-level reinforcement learning algorithm reassesses and reassigns tasks as needed.

### A. Encoder

The encoder is designed based on the attention-based mechanism by focusing on the failure and similarity of past tasks for feature extraction and task prioritization to assist the reinforcement learning algorithm and the energy consumption estimator in making more accurate decisions.

### B. Similarity Attention-based Encoder (SABE)

The primary objective of the SABE is to extract task features and utilize historical data in the experience replay of the high-level reinforcement learning for prioritizing and labeling tasks. These features are then standardized using Min-Max [26] scaling, ensuring they all have equal importance during subsequent computations.

#### 1) Task Similarity Using Attention

Historical data from the RL's experience replay aids in gauging task similarities based on extracted features. Emphasis is on task ID and tasks with significant discrepancies between required and actual execution times and predicted and actual energy consumption. Cosine similarity is utilized to compute the proximity between the feature vectors of new tasks and those stored in the experience replay. Tasks with high discrepancies receive augmented attention weights. More similar tasks get greater attention. An exponential decay mechanism ensures recent tasks have more influence, given by the $e^{-\lambda_1 t}$, where $\lambda_1$ is the decay rate.

#### 2) Task Classification

A task's complexity score incorporates computational demands $C_d$, similar task historical failure in estimation energy consumption $H_f$, and duration $d$. Tasks are categorized into complexity tiers using a threshold-based [27] classification based on these scores. The complexity score $C_s$ is given by

$$C_s = w_1 C_d + w_2 H_f + w_3 d \tag{3}$$

It should be noted that weighting parameters calibrated to balance various operational aspects are shown by $w_i$ in this paper. In addition, tasks are prioritized based on the ground system priority $p$, deadline adherence $D_a$, duration $d$, and differences in required execution time and actual execution time for similar tasks $\delta t_e$ by

$$P_s = w_4 p + w_5 d + w_6 \delta t_e + w_7 D_a \tag{4}$$

Using the TOPSIS method [28], tasks are scored and then classified into complexity and priority tiers, 1-5, based on thresholds.

### C. Task Energy Consumption Estimator

To accurately predict energy consumption before task execution, our methodology leverages an MLP network [29], informed by outputs from the encoder and the high-level experience replay including actual energy consumption of previous tasks.

The estimator utilizes $C_s$, which includes feedback on previous task assumption failures, along with task-specific feature vectors from the encoder. This approach gives more weight to tasks with greater complexity in energy estimation. After each step, we compare predicted energy consumption to actual consumption. Any consistent discrepancies observed serve as a trigger for model re-training.

MLP contains two hidden layers containing 128 and 64 neurons, employing ReLU activation. The model is trained with a Mean Squared Error (MSE) loss and an Adam optimizer [30] at a learning rate of 0.001.

In addition to the regular estimation, we incorporate a 5% energy safety buffer at the start of the process. This safety margin decreases based on the task estimator's error and complexity score. This approach ensures that there is an energy buffer to accommodate unexpected variables or minor inefficiencies that may occur.

### D. Hierarchical Reinforcement Learning Framework

To address the CubeSat task scheduling challenge, we propose a HierRL framework with two primary layers: a high-level policy for global task distribution based on broader constraints, and a low-level policy for real-time monitoring and system adjustment. This lower-level policy handles tasks such as tracking battery levels, meeting deadlines, and monitoring health status, taking corrective actions in the event of failures.

*1) High-Level Task Assigning*

The high-level policy is pivotal in task allocation, utilizing the estimator output to speed up convergence and make informed decisions. Specifically, this policy assigns the most suitable task to each CubeSat based on the current state of the system.

The state of the system consists of the state of the CubeSat includes remind storage and computational resource, remind energy levels of each CubeSat, the temperature of each CubeSat, CubeSats' orientation, required time for each task $t_r$, and the queue of tasks waiting for execution ($P_s$, computational requirement, OW, location, status of processed and estimated energy consumption from estimator).

Actions include assigning and skipping a task, assigning a new task from the task queue to a specific CubeSat and Skip Task, deciding not to assign a task to any CubeSat, and keeping it in the queue for later assignment.

The reward function for $C_i$ at discrete time $t$ is articulated as:

$$R_i(t) = P_s \times w_9 R_{a_i}(t) + w_{10} R_{p_i}(t) + w_{11} R_{e_i}(t) + w_{12} R_{c_i}(t) \tag{5}$$

$R_{e_i}(t) = -\alpha \, \Delta\theta_i(t)$, the observational efficiency, quantifies the CubeSat's efficiency in orienting itself relative to the observational target, where $\Delta\theta_i$ represents the angular difference between the current orientation and the desired orientation of $C_i$ and $\alpha$ is constant that weighs the energy and time costs of reorientation based on the CubeSat's specifications. $R_{c_i}(t) = t_f - (t + \Delta t_i(t))$ is known as OW constraint, which rewards CubeSats that are optimally positioned to complete the observation within the designated window. $t_f$ is the upper limit of the observation window, and $\Delta t_i$ is the time required for $C_i$ to reorient and perform the observation task. In addition, $R_{p_i}$ penalizes excessive energy usage, thereby fostering energy-efficient operations.

The penalty term $R_{P_i} = \sum_{j=1}^{5} R_{P_{i,j}}$ of the reward function is meticulously crafted to penalize the CubeSat for various undesirable actions or states. A prime consideration is energy overconsumption. When a CubeSat's energy consumption surpasses its predefined capacity, a penalty ensues. Given $E_{c_i}(t) = \frac{E_{e_i}}{E_{\max}}$ be representing the normalized energy consumed by $C_i$ for the task, where $E_{e_i}$ is energy consumed, and $E_{\max}$ is the maximum energy capacity of the CubeSat, when $E_{c_i} > 1$, the penalty is quantified as $R_{P_{i,1}} = \beta(E_{c_i}(t) - 1)$. In other instances, it defaults to $E_{c_i}(t)$.

Moreover, for the precision-timed completion of tasks, penalties are imposed when the CubeSat doesn't adhere to set deadlines. If the realized completion time $T_c$ overshoots the task's deadline $D_a$, a stringent penalty, encapsulated as $R_{P_{i,2}}$ is levied.

The energy efficiency penalty is another paramount consideration. Any deviation between the expected energy requirement for a task $E_{e_i}$ and the actual energy consumed $E_{a_i}$ is penalized. The energy deviation is expressed as $\delta E_i(t) = E_{a_i}(t) - E_{e_i}(t)$. The corresponding penalty, aiming to ensure energy-efficient operations, is articulated as $R_{P_{i,3}} = -\lambda_2 E_{e_i}(t) \delta E_i(t)$.

Furthermore, the economical use of computational resources is a keystone in CubeSat operations. In case the computational load $\xi(t)$ exceed its computational resource threshold $\xi_{\max}$, it incurs a penalty delineated by $R_{P_{i,4}}$. Additionally, in the quest to prevent redundant task initiations, penalties are instituted. A CubeSat embarking on a task already in progress by another unit is met with a standard penalty, designated as $R_{P_{i,5}}$.

*Learning Paradigm in High-Level Policy*

In tackling the task scheduling problem within a CubeSat swarm, we have opted for the Multi-Agent Deep Deterministic Policy Gradients (MADDPG) method. The decision is rooted in MADDPG's demonstrated efficacy in maintaining stability within dynamic environments, as well as its aptness for handling scenarios involving multiple agents. Please find the detailed formulation of the actor and critic networks in [31].

In this work, the actor-network is a feed-forward neural network with two layers comprising [256,128] neurons for each actor. Critic-Network has a more complex architecture with three layers of [256,128,64] neurons. The activation function for the output layer is SoftMax, while the hidden layers employ Rectified Linear Unit (ReLU) activation functions.

The training process is structured in episodes, with the task scheduling undergoing training through 20,000 episodes, each representing ten complete orbit cycles. To maintain stability, the Critic network is updated at each time step upon receiving a new task or completing a current one, whereas the Actor-network follows less frequent updates to ensure stability.

Significant hyperparameters include learning rates of 0.001 for the Actor and 0.002 for the Critic. A discount factor $\gamma$ of 0.99 is selected to balance immediate and future rewards effectively. The system incorporates an Ornstein-Uhlenbeck noise process [32] in the action selection mechanism to facilitate the exploration-exploitation trade-off. A replay buffer of size $10^6$ is employed, and mini-batch sizes of 128 are used during training. Following each episode, both the target Critic and Actor networks undergo soft updates with a coefficient of $\tau = 0.005$.

*2) Low-Level Controller Design and Mechanics*

The low-level controller acts with a high-level scheduling mechanism, implemented using a multi-agent deep Q-learning algorithm. While the high-level MADDPG agent is primarily concerned with initial task assignments and global system efficiency, the low-level MADQN focuses on reassigning tasks when unforeseen circumstances arise, such as CubeSat failure or gross underestimation of resource needs.

States include remaining energy level $e_i$, $E_{e_i}$, CubeSat status: *operational* or *in a failure* state, current task ID, the difference between $e_i$ and $E_{e_i}$, and $C_s$. Moreover, the action space consists of two primary actions: keeping the task ($a = 0$) and reallocating the task ($a = 1$), which involves returning it to the global task queue for potential reassignment to another CubeSat.

For the energy-based component of the reward, when the remaining energy surpasses the estimated energy requirement for the task a positive reward is conferred if the decision is to keep the task. Conversely, a penalty is introduced for unnecessary task reassignments. In situations where the energy is insufficient, a significant penalty is enforced to underline the criticality of energy constraints. Mathematically, the energy-based reward $r_e$ is defined by

$$r_e = \begin{cases} \lambda & if \quad e_i > E_{e_i}(1 + C_s/5) \text{ and } a = 0 \\ -\lambda & if \quad e_i > E_{e_i}(1 + C_s/5) \text{ and } a = 1 \\ \ell & if \quad e_i \leq E_{e_i}(1 + C_s/5) \text{ and } a = 0 \end{cases} \quad (6)$$

where $\lambda$ and $\ell$ are constants that determine the magnitude of rewards and penalties.

Moving to the deadline-based reward, if there is ample time to complete the task (adjusted for its priority), the reward mechanism incentivizes retaining the task. However, unwarranted task reassignments in such situations are penalized. When time constraints are tight, a notable penalty is applied to emphasize the importance of task deadlines. Formally, the deadline-based reward is expressed as

$$r_d = \begin{cases} \phi & if \quad D_a > t_r(1 + P_s/10) \text{ and } a = 0 \\ -\phi & if \quad D_a > t_r(1 + P_s/10) \text{ and } a = 1 \\ \psi & if \quad D_a \leq t_r(1 + P_s/10) \text{ and } a = 0 \end{cases} \quad (7)$$

where $\phi$ and $\psi$ are constants dictating the reward and penalty magnitudes, respectively.

Furthermore, to account for the operational status of the CubeSat, a failure penalty is incorporated. In the unfortunate event of a CubeSat failure, a stringent penalty is applied, highlighting the gravity of such an event. This penalty, denoted as $r_f$, is given by

$$r_f = \begin{cases} -\kappa & if \quad \text{CubeSat is failed} \\ 0 & \text{otherwise} \end{cases} \quad (8)$$

with $\kappa$ representing a significant constant value. Aggregating these components, the comprehensive reward of the low-level controller s expressed as $r_{Low-Level} = r_e + r_d + r_f$.

*Training and Learning Paradigm in Low-Level Policy*

The architecture for each agent features two hidden layers, each with 64 neurons activated by ReLU functions. The output layer corresponds to the dimensions of the low-level action space, comprising either "*Reassigning Task*" or "*Keeping Task*", and is activated by a linear function. Further details and the formulation process can be referenced in [33].

The low-level task reassignment is simulated when a CubeSat either failed or is deemed unsuitable for task execution. The DQN algorithm is trained over 20,000 episodes, each characterized by random failure patterns and energy fluctuations. The Q-network is updated at the end of each episode, and experiences are stored in a replay buffer for off-policy learning. The learning rate is set at 0.001, and a decay factor of 0.95 is chosen for the discount rate $\gamma$. The replay buffer size is set at $5 \times 10^5$, and mini-batches of 64 are employed for training. Epsilon begins at 1.0 and decays to 0.01 over 50,000 steps.

*3) Direct Feedback Mechanism*

Feedback from the low-level agent refines the high-level decision process in CubeSat task scheduling. This is achieved by quantifying deviations in task metrics:

$$F = w_{13}(D_a - t - t_r) + w_{14}(e_i - E_{e_i}) + w_{15}N_r \quad (9)$$

where $N_r$ counts task reallocations.

## IV. RESULT AND EVALUATION

This section furnishes a meticulous empirical evaluation of the devised HierRL algorithm for CubeSat task scheduling. Comparative performance analyses are conducted with existing benchmarks, the random policy task assignment algorithm [34], and the MADDPG algorithm. The experiments centered on a constellation of 3, 4, and 5 CubeSats, each tasked with efficiently scheduling observational tasks. The scenarios encompass varying task quantities, specifically 100, 150, and 200 tasks, to evaluate algorithm performance under different workloads.

### A. Experimental Setup

We conducted a series of numerical simulations using MATLAB 2023 to evaluate the proposed scheduling network thoroughly. Given the lack of a standard benchmark for satellite scheduling, we developed a custom scenario generator. This program selects random ground positions and determines the available observation windows for each satellite using the Systems Tool Kit (STK) software.

### B. Results

*1) Evaluation of the algorithm*

Our initial focus was on evaluating the feasibility of the encoder and estimator components. During this phase, we examined the convergence of the task's score under different configurations. These configurations included three combinations of encoder attention weights and estimators. Our observations indicated that higher attention weights led to improved task prioritization, and specific configurations exhibited enhanced learning effects.

During the initial stages of training, we noticed fluctuations in reward feedback. This is attributable to the model's exploration phase, which tries to understand the environment and optimal policy. As training progressed, the Actor-Critic models steadily converged to an optimal policy. Given the vital role of the low-level controller in handling unforeseen challenges, we evaluated its efficiency in reassigning tasks. During simulations where a CubeSat faced unexpected failures or severe energy deviations, the low-level DQN exhibited impressive resilience. The Q-values exhibited stabilization over time. This underscores the effectiveness of the low-level DQN in reallocating tasks when deemed necessary.

In the preliminary stages of training, the estimator's performance was not optimal, primarily due to a scarcity of data. This limitation slightly impeded the quality of our reward function. However, as training progressed and with each successive episode, the estimator's proficiency improved significantly. Fig. 2 shows the mean cumulative rewards in the training phase of the proposed algorithm comparison MADDPG and HierRL without using an encoder and estimator. Our enhanced HierRL framework, integrated with an encoder and estimator, demonstrated rapid and efficient convergence during training. Initially, there was a minor computational overhead due to the encoder's task prioritization and the estimator's energy prediction functionalities. However, as training advanced, these components proved invaluable, steering the system towards optimal decisions faster than traditional algorithms.

The proposed HierRL (with encoder and estimator) converges fastest, benefiting from the encoder's task prioritization and the estimator's energy prediction. Conversely, HierRL (without encoder and estimator) has a slower convergence rate, and MADDPG converges slowest. By convergence's end, the proposed HierRL achieves the highest reward, surpassing its counterpart and MADDPG.

### C. Comparative Analysis with Existing Algorithms

To ascertain the superiority of our proposed HierRL methodology, we compared it with the task scheduling algorithm using MADDPG and random scheduling, assessing metrics like task completion rate and adaptability.

#### 1) Average task success

Fig. 3 highlights the average task success count achieved by the three methodologies (HierRL, random task scheduling, and MADDPG) under examination. Conclusively, HierRL trims the make span by a minimum of 10% compared to the MADDPG, and by at least 15% relative to the Random scheduling, maintaining identical task and CubeSat configurations.

#### 2) Scalability Performance Analysis:

Fig. 4 displays the make span comparison for the three algorithms under varying tasks and CubeSat counts. The make span, in the context of satellite observational tasks, represents the total time required to complete all tasks.

From the figure, our proposed HierRL consistently outperforms both MADDPG and random scheduling across all scenarios. Notably, as the task count rises relative to the number of CubeSats (e.g., 3 CubeSats with 200 tasks), the efficiency of HierRL becomes even more pronounced. This can be attributed to the hierarchical structure and the safety low-level DQN, which reallocates tasks effectively during CubeSat failures and lets the high-level decide globally and divide the responsibility of each level.

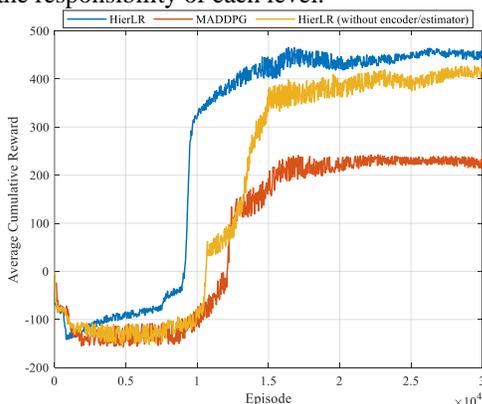

Figure 2 the cumulative reward function of proposed HierRL, HierRL without estimator/encoder, and MADDPG.

In scenarios with a higher CubeSat count relative to tasks (e.g., 5 CubeSats with 100 tasks), the make span difference between HierRL and MADDPG narrows. This suggests that the advantage of the HierRL optimization diminishes as resources (CubeSats) become more abundant. Nonetheless, HierRL still maintains a lead, emphasizing its robustness and adaptability. Random scheduling, lacking any adaptability or learning, consistently lags, showcasing the importance of intelligent scheduling in CubeSat constellations.

### V. CONCLUSION

The empirical evaluation validates the proposed hierarchical reinforcement learning algorithm's effectiveness, efficiency, and robustness for CubeSat task scheduling. Enhanced by advanced features using task scoring using attention mechanism and energy consumption estimator, our model exhibits resilience against network failures and adapts to evolving operational parameters. Therefore, our methodology signifies a groundbreaking advancement in CubeSat swarm task scheduling and resource allocation, warranting further exploration and real-world deployment. Continuous improvement mechanisms, feedback loops, and regular model refinements ensure the system remains cutting-edge and relevant to emerging challenges.

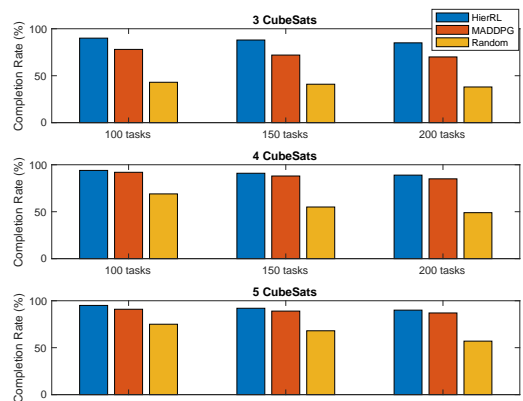

Figure 3 Comparative bar chart displaying the completion of the proposed task scheduling algorithm, random policy, and MADDPG.

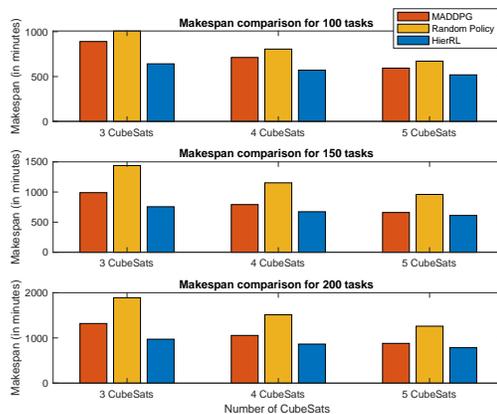

Figure 4 Comparative bar chart displaying the makespan performance of the proposed task scheduling algorithm, random policy, and MADDPG.

REFERENCES

[1] M. Swartwout, "The first one hundred cubesats: A statistical look," *Journal of small Satellites,* vol. 2, no. 2, pp. 213-233, 2013.

[2] M. A. Alandihallaj and M. R. Emami, "Multiple-payload fractionated spacecraft for earth observation," *Acta Astronautica,* vol. 191, pp. 451-471, 2022.

[3] C. Frost, A. Butt, and D. Silva, "Challenges and opportunities for autonomous systems in space," in *Frontiers of Engineering: Reports on Leading-Edge Engineering from the 2010 Symposium*, 2010.

[4] M. Tipaldi and L. Glielmo, "A survey on model-based mission planning and execution for autonomous spacecraft," *IEEE Systems Journal,* vol. 12, no. 4, pp. 3893-3905, 2017.

[5] S. Bandyopadhyay, R. Foust, G. P. Subramanian, S.-J. Chung, and F. Y. Hadaegh, "Review of formation flying and constellation missions using nanosatellites," *Journal of Spacecraft and Rockets,* vol. 53, no. 3, pp. 567-578, 2016.

[6] Y. Chen, A. M. Gillespie, M. W. Monaghan, M. J. Sampson, and R. F. Hodson, "On component reliability and system reliability for space missions," in *2012 IEEE International Reliability Physics Symposium (IRPS)*, 2012: IEEE, pp. 4B. 2.1-4B. 2.8.

[7] M. A. Alandihallaj and M. R. Emami, "Satellite replacement and task reallocation for multiple-payload fractionated Earth observation mission," *Acta Astronautica,* vol. 196, pp. 157-175, 2022.

[8] M. Ghallab, D. Nau, and P. Traverso, *Automated Planning: theory and practice*. Elsevier, 2004.

[9] C. A. Rigo, L. O. Seman, E. Camponogara, E. Morsch Filho, and E. A. Bezerra, "A nanosatellite task scheduling framework to improve mission value using fuzzy constraints," *Expert Systems with Applications,* vol. 175, p. 114784, 2021.

[10] Y. Huang, Z. Mu, S. Wu, B. Cui, and Y. Duan, "Revising the observation satellite scheduling problem based on deep reinforcement learning," *Remote Sensing,* vol. 13, no. 12, p. 2377, 2021.

[11] D. Zhang, X. Han, and C. Deng, "Review on the research and practice of deep learning and reinforcement learning in smart grids," *CSEE Journal of Power and Energy Systems,* vol. 4, no. 3, pp. 362-370, 2018.

[12] R. S. Sutton and A. G. Barto, *Reinforcement learning: An introduction*. MIT press, 2018.

[13] D. L. Poole and A. K. Mackworth, *Artificial Intelligence: foundations of computational agents*. Cambridge University Press, 2010.

[14] Z. Liu, Q. Liu, L. Wang, W. Xu, and Z. Zhou, "Task-level decision-making for dynamic and stochastic human-robot collaboration based on dual agents deep reinforcement learning," *The International Journal of Advanced Manufacturing Technology,* vol. 115, no. 11-12, pp. 3533-3552, 2021.

[15] K. Li, K. Zhang, Z. Zhang, Z. Liu, S. Hua, and J. He, "A uav maneuver decision-making algorithm for autonomous airdrop based on deep reinforcement learning," *Sensors,* vol. 21, no. 6, p. 2233, 2021.

[16] W. J. Wolfe and S. E. Sorensen, "Three scheduling algorithms applied to the earth observing systems domain," *Management Science,* vol. 46, no. 1, pp. 148-166, 2000.

[17] W. Haijiao, Y. Zhen, Z. Wugen, and L. Dalin, "Online scheduling of image satellites based on neural networks and deep reinforcement learning," *Chinese Journal of Aeronautics,* vol. 32, no. 4, pp. 1011-1019, 2019.

[18] A. J. Kleywegt and J. D. Papastavrou, "The dynamic and stochastic knapsack problem," *Operations research,* vol. 46, no. 1, pp. 17-35, 1998.

[19] M. Lemaître, G. Verfaillie, F. Jouhaud, J.-M. Lachiver, and N. Bataille, "Selecting and scheduling observations of agile satellites," *Aerospace Science and Technology,* vol. 6, no. 5, pp. 367-381, 2002.

[20] X. Zhao, Z. Wang, and G. Zheng, "Two-phase neural combinatorial optimization with reinforcement learning for agile satellite scheduling," *Journal of Aerospace Information Systems,* vol. 17, no. 7, pp. 346-357, 2020.

[21] L. Wei, Y. Chen, M. Chen, and Y. Chen, "Deep reinforcement learning and parameter transfer based approach for the multi-objective agile earth observation satellite scheduling problem," *Applied Soft Computing,* vol. 110, p. 107607, 2021.

[22] W. Chong, L. Jun, J. Ning, W. Jun, and C. Hao, "A distributed cooperative dynamic task planning algorithm for multiple satellites based on multi-agent hybrid learning," *Chinese Journal of Aeronautics,* vol. 24, no. 4, pp. 493-505, 2011.

[23] S. Li, R. Wang, M. Tang, and C. Zhang, "Hierarchical reinforcement learning with advantage-based auxiliary rewards," *Advances in Neural Information Processing Systems,* vol. 32, 2019.

[24] H. D. Curtis, *Orbital mechanics for engineering students*. Butterworth-Heinemann, 2013.

[25] K. D. Routley, "A markov game model for valuing player actions in ice hockey," 2015.

[26] B. Radunovic and J.-Y. Le Boudec, "A unified framework for max-min and min-max fairness with applications," *IEEE/ACM Transactions on networking,* vol. 15, no. 5, pp. 1073-1083, 2007.

[27] L. Mohammadi and S. van de Geer, "On threshold-based classification rules," *Lecture Notes-Monograph Series,* pp. 261-280, 2003.

[28] M. Behzadian, S. K. Otaghsara, M. Yazdani, and J. Ignatius, "A state-of the-art survey of TOPSIS applications," *Expert Systems with applications,* vol. 39, no. 17, pp. 13051-13069, 2012.

[29] S. Osowski, K. Siwek, and T. Markiewicz, "MLP and SVM networks-a comparative study," in *Proceedings of the 6th Nordic Signal Processing Symposium, 2004. NORSIG 2004.*, 2004: IEEE, pp. 37-40.

[30] Z. Zhang, "Improved adam optimizer for deep neural networks," in *2018 IEEE/ACM 26th international symposium on quality of service (IWQoS)*, 2018: Ieee, pp. 1-2.

[31] R. Lowe, Y. I. Wu, A. Tamar, J. Harb, O. Pieter Abbeel, and I. Mordatch, "Multi-agent actor-critic for mixed cooperative-competitive environments," *Advances in neural information processing systems,* vol. 30, 2017.

[32] P. Cheridito, H. Kawaguchi, and M. Maejima, "Fractional ornstein-uhlenbeck processes," 2003.

[33] M. Egorov, "Multi-agent deep reinforcement learning," *CS231n: convolutional neural networks for visual recognition,* pp. 1-8, 2016.

[34] F. Semchedine, L. Bouallouche-Medjkoune, and D. Aissani, "Task assignment policies in distributed server systems: A survey," *Journal of network and Computer Applications,* vol. 34, no. 4, pp. 1123-1130, 2011.